\documentclass[sigconf,letterpaper,9pt]{acmart}

\makeatletter                   %1
\def\mdseries@tt{m}             %1
\makeatother                    %1
\usepackage[plain]{fancyref}
\usepackage[draft=true]{minted} %2
\usepackage{color}
\usepackage{hyperref}           %5
\hypersetup{
    colorlinks=true,
    linkcolor=blue,
    filecolor=red,
    urlcolor=magenta,
    breaklinks=true,            %3
}
\usepackage{breakurl}           %3

\usepackage{xcolor}
\usepackage{booktabs}

\usepackage{listings}
\usepackage[]{graphicx}
\usepackage{subcaption}
\usepackage{sistyle}
\SIthousandsep{,}
\usepackage{enumitem}
\usepackage{changes}
\usepackage{xspace}
\usepackage{hyperref}

\providecommand{\xb}{\mathbf{x}}
\providecommand{\RR}{\mathbb{R}}
\newcommand{\clarify}{{\tt Clarify}\xspace}

\newif\ifdraft \drafttrue

\AtBeginDocument{%
  \providecommand\BibTeX{{%
    \normalfont B\kern-0.5em{\scshape i\kern-0.25em b}\kern-0.8em\TeX}}}

\setcopyright{none}
\renewcommand\footnotetextcopyrightpermission[1]{} % removes footnote with conference information in first column
\begin{document}

% \copyrightyear{2021}
% \acmYear{2021}
\acmConference[KDD '21]{Proceedings of the 27th ACM SIGKDD Conference on Knowledge Discovery and Data Mining}{August 14--18, 2021}{Virtual Event, Singapore}
% \acmBooktitle{Proceedings of the 27th ACM SIGKDD Conference on Knowledge Discovery and Data Mining (KDD '21), August 14--18, 2021, Virtual Event, Singapore}\acmDOI{10.1145/3447548.3467177}
% \acmISBN{978-1-4503-8332-5/21/08}
% \fancyhead{}

\title{Amazon SageMaker Clarify: Machine Learning Bias Detection and Explainability in the Cloud}

\author{Michaela Hardt$^1$, Xiaoguang Chen, Xiaoyi Cheng, Michele Donini, Jason Gelman, \mbox{Satish Gollaprolu,
John He, Pedro Larroy, Xinyu Liu,} Nick McCarthy, Ashish Rathi, \mbox{Scott Rees, Ankit Siva, ErhYuan Tsai$^2$, Keerthan Vasist,} Pinar Yilmaz, Muhammad Bilal Zafar, \mbox{Sanjiv Das$^3$, Kevin Haas, Tyler Hill}, Krishnaram Kenthapadi}
\affiliation{%
 \institution{Amazon Web Services}
 %\city{East Palo Alto, CA}
 \country{ %USA\\
{\small \it Appeared in the proceedings of the 27th ACM SIGKDD Conference on Knowledge Discovery and Data Mining (KDD 2021)}
 }
 }
 
 \thanks{$^1$Correspondence to: Michaela Hardt, <milaha@amazon.com>.}  %
\thanks{$^2$Work done while at Amazon Web Services.}  %
 \thanks{$^3$Santa Clara University, CA, USA}

\renewcommand{\shortauthors}{M. Hardt et al.}

\begin{abstract}
Understanding the predictions made by machine learning (ML) models and their potential biases remains a challenging and labor-intensive task that depends on the application, the dataset, and the specific model. We present Amazon SageMaker Clarify, an explainability feature for Amazon SageMaker that launched in December 2020, providing insights into data and ML models by identifying biases and explaining predictions. It is deeply integrated into Amazon SageMaker, a fully managed service that enables data scientists and developers to build, train, and deploy ML models at any scale. Clarify supports bias detection and feature importance computation across the ML lifecycle, during data preparation, model evaluation, and post-deployment monitoring. We outline the desiderata derived from customer input, the modular architecture, and the methodology for bias and explanation computations. Further, we describe the technical challenges encountered and the tradeoffs we had to make. For illustration, we discuss two customer use cases. We present our deployment results including qualitative customer feedback and a quantitative evaluation. Finally, we summarize lessons learned, and discuss best practices for the successful adoption of fairness and explanation tools in practice.

\end{abstract}


% \begin{CCSXML}
% <ccs2012>
%    <concept>
%        <concept_id>10010147.10010257</concept_id>
%        <concept_desc>Computing methodologies~Machine learning</concept_desc>
%        <concept_significance>500</concept_significance>
%        </concept>
%    <concept>
%        <concept_id>10010147.10010919.10010172</concept_id>
%        <concept_desc>Computing methodologies~Distributed algorithms</concept_desc>
%        <concept_significance>100</concept_significance>
%        </concept>
%    <concept>
%        <concept_id>10011007.10011006</concept_id>
%        <concept_desc>Software and its engineering~Software notations and tools</concept_desc>
%        <concept_significance>100</concept_significance>
%        </concept>
%  </ccs2012>
% \end{CCSXML}

% \ccsdesc[500]{Computing methodologies~Machine learning}
% \ccsdesc[100]{Computing methodologies~Distributed algorithms}
% \ccsdesc[100]{Software and its engineering~Software notations and tools}

\maketitle

\section{Introduction}\label{sec:intro}
Machine learning (ML) models and data-driven systems are increasingly used to assist in  decision-making across domains such as financial services, healthcare, education, and human resources. Benefits of using ML include improved accuracy, increased productivity, and cost savings. The increasing adoption of ML is the result of multiple factors, most notably ubiquitous connectivity, the ability to collect, aggregate, and process large amounts of data using cloud computing, and improved access to increasingly sophisticated ML models. In high-stakes settings,  tools for bias and explainability in the ML lifecycle are of particular importance.

{\noindent \em Regulatory}~\cite{wh2014bigdata,wh2016bigdata}: Many ML scenarios require an understanding of why the ML model made a specific prediction or  whether its prediction was biased. %
Recently, policymakers, regulators, and advocates 
have expressed concerns about the potentially discriminatory impact of ML and data-driven systems%
, for example, due to inadvertent encoding of bias into automated decisions. Informally, biases can be viewed as imbalances in the training data or the prediction behavior of the model across different groups, such as age or income bracket.

{\noindent \em Business}~\cite{ftc2016bigdata,goodman2017european}: The adoption of AI systems in regulated domains requires %
explanations of %
how models make predictions. 
Model explainability is particularly important in industries with reliability, safety, and compliance requirements, such as financial services, human resources, healthcare, and automated transportation. Using a financial example of lending applications, loan officers, customer service representatives, compliance officers, and end users often require explanations detailing why a model made certain predictions.

{\noindent \em Data Science}: Data scientists and ML engineers need tools %
to debug and improve ML models, to ensure a diverse and unbiased dataset during feature engineering, to determine whether a model is making inferences based on noisy or irrelevant features, and to understand the limitations and failure modes of their model~\cite{slack2021defuse}. %

While there are several open-source tools for fairness and explainability, we learned from our customers that developers often find it tedious to incorporate these tools across the ML workflow, as this requires a sequence of complex, manual steps: (1) Data scientists and ML developers first have to translate internal compliance requirements into code that can measure bias in their data and ML models. To do this, they often must exit their ML workflow, and write custom code to piece together open source packages. (2) Next, they need to test datasets and models for bias across all steps of the ML workflow and write code for this integration. (3) Once the tests are complete, data scientists and ML developers create reports showing the factors that contribute to the model prediction by manually collecting a series of screenshots for internal risk and compliance teams, auditors, and external regulators. (4) Finally, after the models are in production, data scientists and ML developers must periodically run through a series of manual tests to ensure that no bias is introduced as a result of a drift in the data distribution (e.g., changes in income for a given geographic region). Similar challenges have also been pointed out in prior studies~\cite{holstein2019improving,mitchell2019model,bhatt2020explainable}.

{\noindent \bf Our contributions}: To address the need for generating bias and explainability insights across the ML workflow, we present Amazon SageMaker \clarify\footnote{\url{https://aws.amazon.com/sagemaker/clarify}}, a new functionality in Amazon SageMaker~\cite{sagemaker} that helps detect bias in data and models, helps explain predictions made by models, and generates associated reports. 
We describe the design considerations and technical architecture of \clarify, and its integration across the ML lifecycle (\S\ref{sec:design}).
We then discuss the methodology for bias and explainability computations (\S\ref{sec:methods}). We implemented several measures of bias in \clarify, considering the dataset (pre-training) and the model predictions (post-training), to support various use cases of SageMaker's customers.
We integrated pre-training bias detection with the data preparation functionality and post-training bias detection with the model evaluation functionality in SageMaker. These biases can be monitored continuously for deployed models and alerts can notify users about changes.
For model explainability, \clarify offers a scalable implementation of SHAP (SHapley Additive exPlanations)~\cite{lundberg_unified_2017}, based on the concept of the Shapley value~\cite{shapley_value_1952}. 
In particular, we implemented the KernelSHAP~\cite{lundberg_unified_2017} variant as it is model agnostic, so that \clarify can be applied to the broad range of models trained by SageMaker's customers. For deployed models,  \clarify monitors for changes in feature importance and issues alerts.

We discuss the implementation and technical challenges in \S\ref{sec:development} and illustrate the graphical representation in SageMaker Studio in a case study in \S\ref{sec:case_study}.
In \S\ref{sec:results} we present the deployment results, including a quantitative evaluation of the scalability of our system, a report on usage across industries, qualitative customer feedback on usability, and two customer use cases.
Finally, we summarize lessons learned in \S\ref{sec:lessons} and conclude in \S\ref{sec:conclusions}.

\vspace{-2mm}
\section{Related Work}\label{sec:related_work}
\clarify offers bias detection and explainability of a model. %
Both are active research areas that we review briefly below.   

{\it Bias.}
Proposed bias metrics for machine learning include individual fairness~\cite{dwork-fairness,NEURIPS2019_0e1feae5} as well as group fairness. Group fairness considers different aspects of model quality (e.g., calibration~\cite{kleinberg2016inherent}, error rates~\cite{hardt_equality_2016}), %
and prediction tasks (e.g., ranking~\cite{rankingfair} or classification ~\cite{fair-classification,donini_empirical_2018}). 
Techniques based on causality~\citep{pearl_causal_2009} provide promising alternatives at the individual~\cite{counterfactual_failr} and group level~\cite{causal_fair}.
For a thorough overview see ~\cite{barocas_fairness_2019,mehrabi_survey_2019}. %

Techniques to mitigate bias either change the model, its optimization,  (e.g.,~\cite{agarwal_reductions_2018, adv_fairness, perrone2020fair}), or post-process its predictions (e.g.,~\cite{hardt_equality_2016,geyik2019fairness}).
There are several open source repositories that offer bias management approaches, including fairness-comparison\footnote{\url{https://github.com/algofairness/fairness-comparison}}, Aequitas\footnote{\url{https://github.com/dssg/aequitas}}, Themis\footnote{\url{https://github.com/LASER-UMASS/Themis}}, responsibly\footnote{\url{https://github.com/ResponsiblyAI/responsibly}}, and LiFT~\cite{vasudevan20lift}. Moreover,
IBM's AI Fairness 360\footnote{\url{https://aif360.mybluemix.net/}}, Microsoft's Fairlearn\footnote{\url{https://fairlearn.github.io/}}, and TensorFlow's What-If-Tool~\cite{what-if} are products built around bias metrics and mitigation strategies.

{\it Explainabilty.}
The approaches for model explainability can be dissected in various ways. For a comprehensive survey, see e.g.~\cite{guidotti2018survey}.
We distinguish \textit{local vs. global explainability} approaches. Local approaches~\cite{ribeiro2016should,lundberg_explainable_2018} explain model behavior in the neighborhood of a given instance. Global explainability methods aim to explain the behavior of the model at a higher-level (e.g.~\cite{lakkaraju_faithful_2019}). 
Explainability methods can also be divided into \textit{model-agnostic} (e.g., \cite{lundberg_explainable_2018}) and  \textit{model-specific} approaches which leverage internals of models such as gradients~(e.g., \cite{simonyan_deep_2014,sundararajan_axiomatic_2017}). %
There is growing impetus to base explanations on counterfactuals \citep{barocas_hidden_2020,janzing_feature_2019}. At the same time concern about overtrusting explanations grows~\citep{mythos, rudin_stop_2019}.

In comparison to previous work, \clarify is deeply integrated into SageMaker and couples scalable fairness and explainability analyses. For explainability, its implementation of SHAP offers  model-agnostic feature importance at the local and global level.

\section{Design, Architecture, Integration}\label{sec:design}

\subsection{Design Desiderata}
Conversations with customers in need of detecting biases in  models and explaining their predictions  shaped the following  desiderata.
\begin{itemize}
    \item \textbf{Wide Applicability.} Our offerings should be widely applicable across %
(1) industries (e.g., finance, human resources, and healthcare), (2) use cases (e.g., click-through-rate prediction, fraud detection, and credit risk modeling), (3) prediction tasks (e.g., classification and regression), (4) ML models (e.g., logistic regression, decision trees, deep neural networks), and (5) the ML lifecycle (i.e., from dataset preparation to model training to monitoring deployed models).
    \item \textbf{Ease of Use.} Our customers should not have to be experts in fairness and explainability, or even in ML. They should be able to analyze bias and obtain feature importance within their existing workflow having to write little or no code. 
    \item \textbf{Scalability.} Our methods should scale to both complex models (i.e., billions of parameters) and large datasets. 
\end{itemize}
The latter two desiderata were the top two non-functional requirements requested as customers evaluated the early prototypes. Customers described frustratingly long running times with existing open source techniques, which motivated us to scale these existing solutions to large datasets and models (c.f. \S\ref{sec:scalability}). %
Moreover, feedback from early prototypes included frequent erroneous configurations due to the free-form nature of the configuration file, prompting us to improve usability (c.f. \S\ref{sec:usability}). %

\subsection{Architecture of \clarify}
\begin{figure}[h]
\vspace{-4mm}
\begin{center}
\includegraphics[scale=0.4]{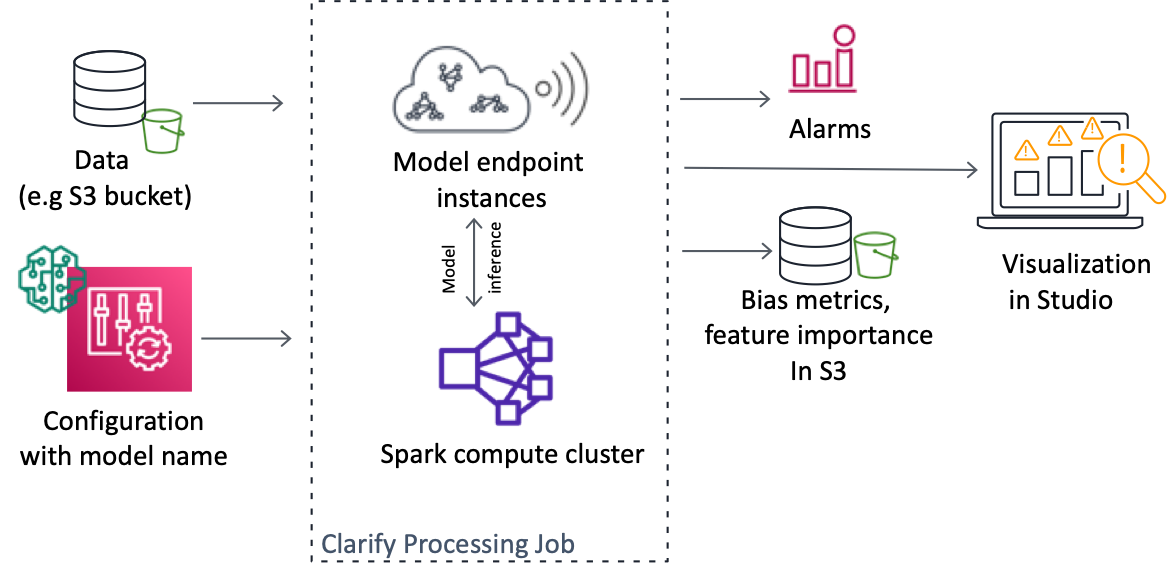}
\vspace{-3mm}
\caption{High-level system architecture of \clarify. A dataset and a configuration file serve as input to a batch processing job that produces bias metrics and feature importance. These outputs are stored in a filesystem, exported as metrics, and visualized in a web-based IDE for ML.
\label{fig:system}}
\end{center}
\vspace{-4mm}
\end{figure}
\clarify is implemented as part of Amazon SageMaker~\cite{sagemaker}, a fully managed service that enables data scientists and developers to build, train, and deploy ML models at any scale. It leverages SageMaker’s processing job APIs, and executes batch processing jobs on a cluster of AWS compute instances in order to process large amounts of data in a scalable manner. 
Each processing job has an associated cluster of fully managed SageMaker compute instances running the specified container image and provisioned specifically for the processing job. 
\clarify provides a first-party container image which bundles efficient and optimized implementations of core libraries for bias and explainability computation together with flexible parsing code to support a variety of formats of datasets and the model input and output. %

Inputs to \clarify are the dataset stored in the S3 object store and a configuration file specifying the bias and explainability metrics to compute and which model to use. At a high level, the processing job completes the following steps: (1) validate inputs and parameters; (2) compute pre-training bias metrics; (3) compute post-training bias-metrics; (4) compute local and global feature attributions for explainability; (5) generate output files.
The job results are saved in the S3 object store and can be consumed in multiple ways: (1) download and programmatically inspect using a script or notebook; (2) visualize and inspect in SageMaker Studio IDE through custom visualizations and tooltips to describe and help interpret the results.

The cluster of compute instances carries out the computation in a distributed manner (c.f. \S\ref{sec:development}). 
To obtain predictions from the model, the \clarify job spins up model serving instances that take inference requests. This is a deliberate choice made to avoid interfering with production traffic to the already deployed models and polluting the performance and quality metrics of the models being monitored in production. %
This resulted in a modular system allowing customers to use our product without any changes to their deployed models; see Figure~\ref{fig:system}. 

{\noindent \bf Security.}
Built on top of existing SageMaker primitives, \clarify supports access control and encryption for the  data and the network. Specifically, roles, virtual private cloud configurations, and encryption keys can be specified.

\subsection{Deep Integration in SageMaker} \label{sec:integration}
 \begin{figure*}[h]
 \vspace{-8mm}
\begin{center}
\includegraphics[scale=0.40]{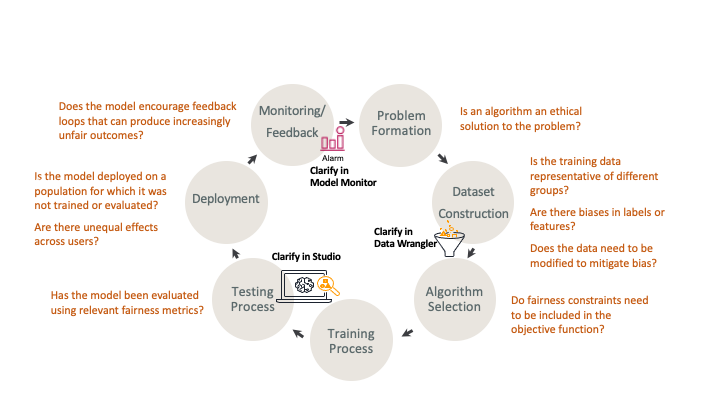} 
\vspace{-4mm}
\caption{The stages in a ML lifecycle from problem formulation to deployment and monitoring with their respective fairness considerations in orange. \clarify can be used at dataset construction and model testing and monitoring to investigate them.
\label{fig:cycle}}
\end{center}
\vspace{-4mm}
\end{figure*}
For generating bias and explainability insights across the ML lifecycle, we integrated \clarify with the following SageMaker components: Data Wrangler to visualize dataset biases
during data preparation, Studio \& Experiments to explore biases and explanations of trained models, and Model Monitor to monitor these metrics.\footnote{Data Wrangler: \url{aws.amazon.com/sagemaker/data-wrangler}; Studio: \url{aws.amazon.com/sagemaker/studio}; Model Monitor: \url{aws.amazon.com/sagemaker/model-monitor}.} 
Figure~\ref{fig:cycle} illustrates the stages during ML lifecycle and the integration points of \clarify within SageMaker.

\section{Methods}\label{sec:methods}

\subsection{Bias} \label{sec:bias}
We implemented several bias metrics to support a wide range of use cases. The selection of the appropriate bias metric for an application requires human judgement. 
Some of the metrics only require the dataset and thus can be computed early in the model lifecycle, that we call \emph{pre-training bias metrics}. They can shed light onto how different groups are represented in the dataset and if there is a difference in label distribution across groups. At this point some biases may require further investigation before training a model. 
Once a model is trained, we can compute \emph{post-training bias metrics} that take into account how its predictions differ across groups.
We introduce the notation before defining the metrics. 

\subsubsection{Setup and Notation}
Without loss of generality, we use $d$ to denote the group that is potentially disadvantaged by the bias and compare it to the reminder of the examples in the dataset $a$ (for advantaged group). %
The selection of the groups often requires domain information%
\footnote{\url{https://plato.stanford.edu/entries/discrimination/}}
and 
application specific considerations.
Denote the number of examples of each group by $n_a, n_d$, with $n$ denoting the total.
We assume that each example in the dataset has an associated binary label $y \in \{0,1\}$ (with $1$ denoting a positive outcome), and that the ML model predicts $\hat{y} \in \{0,1\}$.
While these assumptions seem restrictive, different prediction tasks can be re-framed accordingly. For example, for regression tasks we can select one or more thresholds of interest. For multi-class predictions we can treat various subsets of the classes as positive outcomes.

We denote by  $n^{(0)},n^{(1)}$, respectively, the number of  labels of value 0, 1 that we can further break down by group into $n_a^{(0)}, n_a^{(1)}$ and $n_d^{(0)}, n_d^{(1)}$. 
Analogously, we can count the positive / negative predictions $\hat{n}^{(0)}, \hat{n}^{(1)}$ and break them down by group into $\hat{n}_a^{(0)}, \hat{n}_a^{(1)}$ and $\hat{n}_d^{(0)}, \hat{n}_d^{(1)}$.
These counts can be normalized by group size as $q_a=n_a^{(1)}/n_a$ and $q_d=n_d^{(1)}/n_d$ for the labels and $\hat{q}_a=\hat{n}_a^{(1)}/n_a$ and $\hat{q}_d=\hat{n}_d^{(1)}/n_d$ for the predictions. %
Additionally, we consider the $TP$: true positives; $FP$: false positives; $TN$: true negatives; $FN$: false negatives; and compute them for each group separately.

\subsubsection{Pre-Training Metrics}

\begin{enumerate}[leftmargin=*]
\item[$CI$] Class Imbalance: Bias can stem from an under-representation of the disadvantaged group in the dataset, for example, due to  a sampling bias. We define $CI = \frac{n_a-n_d}{n}$.

\item[$DPL$] Difference in positive proportions in observed labels: We define $DPL=q_a-q_d$, to measures bias in the label distribution. Additional metrics derived from $q_a$ and $q_d$ are defined in \S\ref{sec:appendix_bias_methods}\footnote{\url{https://pages.awscloud.com/rs/112-TZM-766/images/Amazon.AI.Fairness.and.Explainability.Whitepaper.pdf}}.

\item[$CDDL$] Conditional Demographic Disparity in Labels: Can the difference in $q_a$ and $q_d$ be explained by another feature?
An instance of Simpson's paradox\footnote{\url{https://www.britannica.com/topic/Simpsons-paradox}} arose in the case of Berkeley admissions where men were admitted at a higher rate than women %
overall, but for each department women were admitted at a higher rate. In this case women applied more to departments with lower acceptance rates.
Following \cite{wachter_why_2020}, we define $DD=n_d^{(0)}/n^{(0)}-n_d^{(1)}/n^{(1)}$ and compute it across all strata $i$ on a user-supplied attribute that we want to control for, yielding $DD_i$. 
$CDDL = \frac{1}{n} \;\sum_i n_i \cdot DD_i$
where $i$ subscripts each strata and $n_i$ is the number of examples in it. %
\end{enumerate}

\subsubsection{Post-Training Metrics}
\label{sec_posttrain}
\begin{enumerate}[leftmargin=*]
\item[$DPPL$] Difference in positive proportions in predicted labels: $DPPL$ consider the difference in positive predictions: $DPPL = \hat{q}_a-\hat{q}_d$, related is ``statistical parity''~ \cite{berk_fairness_2017, calders_three_2010}.

\item[$DI$] Disparate Impact: Instead of the difference as in $DPPL$ we can consider the ratio:
$
DI =  \hat{q}_d / \hat{q}_a.
$

\item[$DCA$] Difference in Conditional Acceptance / Rejection: $DCA$ considers the ratio of the number of positive labels to that of positive predictions assessing calibration: $DCA=n_a^{(1)}/\hat{n}_a^{(1)}-n_d^{(1)}/\hat{n}_d^{(1)}$. 
Analogously, $DCR$ considers negative labels and predictions.

\item[$AD$] Accuracy Difference: We compare the accuracy, i.e. the fraction of examples for which the predictions is equal to the label, across groups. %
We define $AD = (TP_a + TN_a) / n_a - (TP_d + TN_d) / n_d$.

\item[$RD$] Recall Difference: We compare the recall (the fraction of positive examples that receive a positive prediction) across groups.
$RD = TP_a/{n}_a^{(1)} - TP_d/{n}_d^{(1)}$ (aka ``equal opportunity''~\cite{hardt_equality_2016}).

\item[$DAR$] Difference in Acceptance / Rejection Rates: This metric  considers precision, i.e. the fraction of positive predictions that are correct: $DAR = TP_a / \hat{n}_a^{(1)} - TP_d / \hat{n}_d^{(1)}$.
Analogously, we define $DRR$ for negative predictions.

\item[$TE$] Treatment Equality:  We define $TE = FN_d/FP_d - FN_a/FP_a$ \cite{berk_fairness_2017}.

\item[$CDDPL$] Conditional Demographic Disparity of Predicted Labels: This metric is analogous to the $CDDL$ for predictions.
\end{enumerate}
See \S\ref{sec:appendix_bias_methods} for additional metrics.

\vspace{-3.5mm}
\subsection{Explainabilty} \label{sec:shap}
We  implemented  the KernelSHAP~\cite{lundberg_unified_2017} algorithm due to its good performance~\cite{ribeiro2016should} and wide applicability.

The KernelSHAP algorithm~\cite{lundberg_unified_2017} works as follows:
Given an input $\xb \in \RR^M$ and the model output $f(\xb) \in \RR$, SHAP, building on the Shapley value framework~\citep{shapley_value_1952}, designates the prediction task as a $M$-player cooperative game, with individual features being players and the predicted output $f(\xb)$ being the total payout of the game. 
The framework decides how \textit{important} each player is to the game's outcome.
Specifically, the importance of the $i^{th}$ player $\xb_i$ is defined as the average marginal contribution of the player to a game not containing $\xb_i$. 
Formally, the contribution $\phi_i$ is defined as:
\begin{equation*}
    \phi_i(\mathbf{x}) = \sum_{S} \frac{|S|! (M - |S| - 1)!}{M!} \big( f(\mathbf{x}_S \cup \mathbf{x}_i) - f(\mathbf{x}_S) \big),
\end{equation*}
where the sum is computed over all the player coalitions $S$ not containing $\xb_i$.
The computation involves iterating over $2^M$ feature coalitions, which does not scale to even moderate values of $M$, e.g., 50. To overcome these scalability issues, KernelSHAP uses approximations~\cite{lundberg_unified_2017}.

\subsection{Monitoring bias and explanations}
Measuring bias metrics and feature importance provide insights into the model at time of evaluation. However, a change in data distribution can occur over time, resulting in different bias metrics and feature importance. Our goal is to continuously monitor for such changes and raise alerts if bias metrics or feature importance drift from \textit{reference values} established during an earlier evaluation (before the model launch). Based on a user-specified schedule (e.g., hourly), newly accumulated data is analyzed and compared against the reference values. 

{\noindent \bf Bias.}
Given a bias metric from \S\ref{sec:bias}, we let the user specify a range of reference values $A = (a_\text{{min}}, a_\text{{max}})$ in which the new bias value $b$ computed on  the live data should lie. However, if little data was accumulated simply checking for containment can lead to noisy results. 
To ensure that the conclusions drawn from the observed live data  are statistically significant, we check for overlap of $A$ with the $95\%$ bootstrap confidence interval of $b$~\cite{efron1994introduction}. %

{\noindent \bf Explainability.}
For feature importance we consider the change in ranking of features by their importance. Specifically, we use the Normalized Discounted Cumulative Gain (nDCG) score for comparing the ranking by feature importance of the reference data with that based on the live data. nDCG penalizes changes  further down in the ranking less. Also, swapping the position of two values is penalized relative to the difference of their original importance.

Let $F = [f_1, \ldots, f_{M}]$ denote the list of $M$ features sorted w.r.t. their importance scores in the reference data, and $F' = [f'_1, \ldots, f'_{M}]$ denote the new ranking of the features based on the live data.  We use $a(f)$ to describe a function that returns the feature importance score on the reference data of a feature $f$. Then, nDCG is defined as: $\text{nDCG} = \text{DCG} / \text{iDCG}$, where  $\text{DCG} = \sum_{i=1} ^ {M} a(f'_i) / \text{log}_2(i+1)$ and $\text{iDCG} = \sum_{i=1} ^ {M} a(f_i)/ \text{log}_2(i+1)$.

An nDCG value of 1 indicates no change in the ranking. In our implementation, an nDCG value below 0.90 triggers an alert.

\section{Alternative Design and Methods} \label{sec:alt_designs}
On the architectural side, our design provisions a separate model server which adds latency. 
Alternatively, a design that runs the model locally avoids network traffic. 
We decided to use a model server so that we can re-use the existing SageMaker model hosting primitives.  Our model-agnostic approach enables support of a wide range of models including those that cannot be executed locally. This way we can scale and tune the instance type selection and counts for the processing job and the model serving instances independently based on the dataset and model characteristics. 

For the bias metrics, we considered limiting the number of metrics, to avoid overwhelming users. However, the appropriate bias metric  is application dependent. To support customers from different industries with various use cases, we decided to implement a wide range of metrics (see  \S\ref{sec:bias}).

For explainability we decided to implement  the KernelSHAP~\cite{lundberg_unified_2017} algorithm (see  \S\ref{sec:shap}) rather than alternatives discussed in \S\ref{sec:related_work} for a number of reasons: 
(i) it offers wide \textit{applicability} to interpret predictions of any score-based model;
(ii) it works in a \textit{model-agnostic} manner and does not require access to model internals such as gradients, allowing us to build on top of the the SageMaker hosting primitives;
(iii) it provides attractive theoretical properties associated with the Shapley value framework~\citep{shapley_value_1952};
and
(iv) benchmarking shows that it performs favorably compared to related methods such as LIME~\cite{ribeiro2016should}.
More efficient model-specific implementations like TreeSHAP and DeepSHAP~\cite{lundberg_unified_2017} leverage the model architecture for improved efficiency. By using the model-agnostic KernelSHAP, we traded efficiency for wider applicability. 
In the future we will prioritize support for additional explainability methods based on customer requests.

\section{Development}\label{sec:development}
{\noindent \bf Timeline:}
We started by collecting customer requirements to derive \clarify's goals and design the system (\S\ref{sec:design}). Our first prototype implemented 
 the methods on a single node packaged into a container that could be run with a bare-bone configuration file. After internal tests, we deployed this version for beta customers to solicit feedback.
 This feedback helped us to prioritize scalability (\S\ref{sec:scalability}) and ease of use improvements (\S\ref{sec:usability}).
Afterwards, we expanded support of models and datasets, conducted more end-to-end tests, and obtained additional rounds of customer feedback.
We embarked on the integration across the model lifecycle (\S\ref{sec:integration}), emphasizing the associated UI components and documentation.\footnote{\url{https://docs.aws.amazon.com/sagemaker/latest/dg/clarify-detect-post-training-bias.html}, \url{https://docs.aws.amazon.com/sagemaker/latest/dg/clarify-model-explainability.html}}
Prior to public launch, we set up pipelines and dashboards for health monitoring of our system,
including alarms for too many failed jobs. %
Throughout, multiple reviews took place for the design, usability, security and privacy, and  operational readiness.

{\noindent \bf Implementation Details:}
For wide applicability, we implemented flexible parsers to support various data formats (JSONLines/CSV, with(-out) header) and model signatures, see \S\ref{sec:appendix_predictor_config} for examples.

The first bias metrics were implemented in numpy.\footnote{Our open source implementation is available at \url{https://github.com/aws/amazon-sagemaker-clarify}.} 
For scalability, all subsequent methods were implemented using Spark via SageMaker's Spark Processing job. 
Spark provided us with convenient abstractions (e.g., distributed median computations) and managed the cluster for us. 
Our bias metrics mostly require counts that are easy to distribute. Distributing SHAP only requires one map operation to calculate feature importance for each example followed by an aggregation for global model feature importance.

\subsection{Challenges and Resolutions}\label{sec:dev_challenges}

A challenge to fault-tolerance is that \clarify's offline batch processing job  hits a live model endpoint. This requires a careful implementation to avoid overwhelming the model endpoint with requests. 
To limit the potential impact, \clarify spins up separate model servers and does not interfere with production servers. %
To avoid overwhelming these separate model endpoints, we dynamically configured the batch size through experimentation, starting with the largest possible batch size based on the maximum payload\footnote{\url{https://github.com/boto/botocore/blob/develop/botocore/data/sagemaker-runtime/2017-05-13/service-2.json##L35}} in order to achieve high throughput at the model endpoints.
Furthermore, we relied on the boto clients' retry policies\footnote{\url{https://boto3.amazonaws.com/v1/documentation/api/latest/guide/retries.html}} with exponential back-off. This is necessary to tolerate intermittent inference call failures.
We further expanded the retry mechanism, since not all errors are considered retriable in the boto client.

A key to scalability was to utilize resources well by carefully tuning the number of partitions of the dataset in Spark:
If tasks took very long (we started out with tasks taking an hour) for SHAP computation, failures were costly and utilization during the last few tasks was low. 
If tasks were too small for bias metrics computation, throttling limitations when instantiating a SageMaker Predictor led to job failures.
The solution was to carefully tune the number of tasks separately for explainability and bias computations.

\section{Illustrative Case Study}\label{sec:case_study}
\begin{figure}
\vspace{-4mm}
\begin{center}
  \includegraphics[width=1\linewidth]{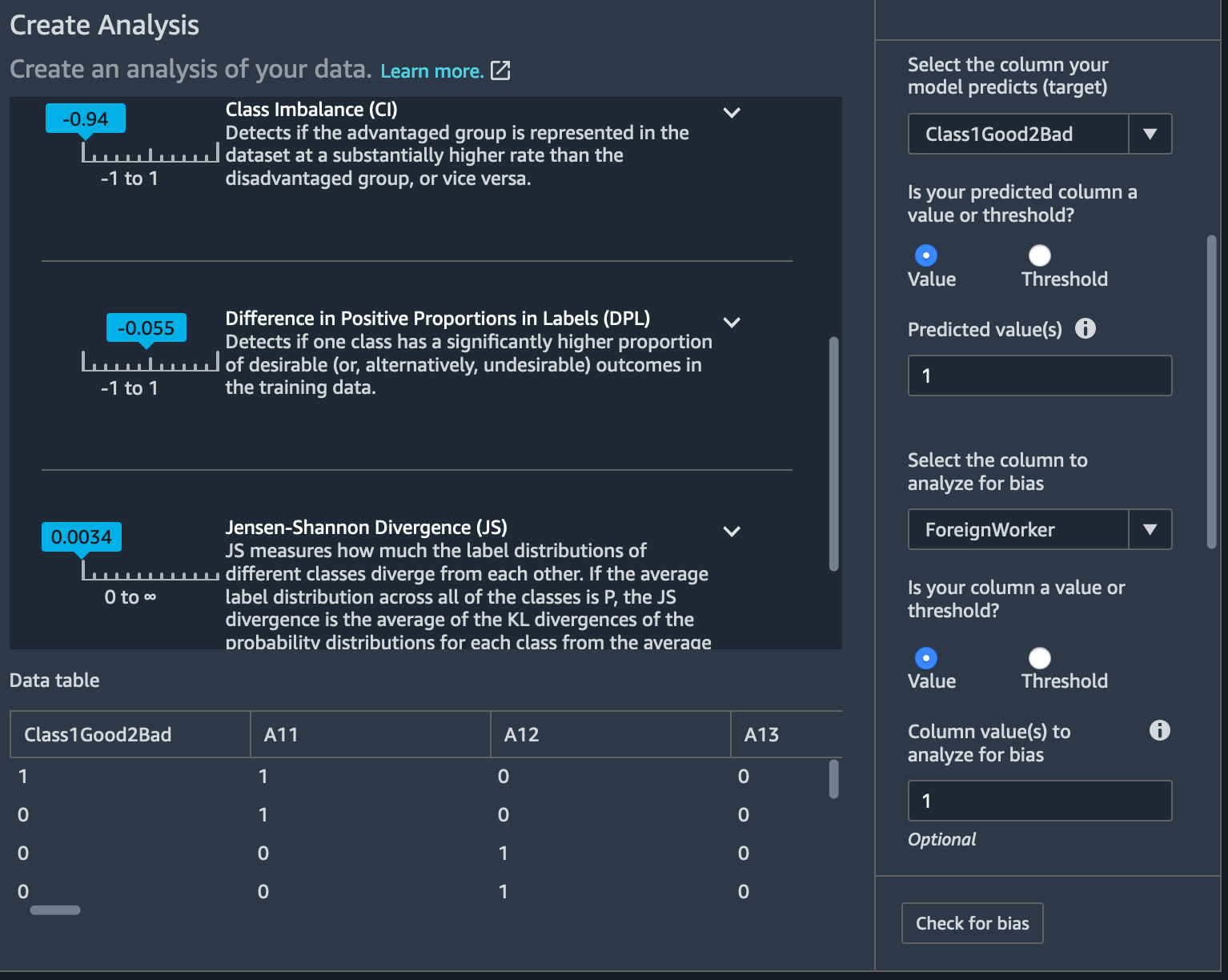}
  \vspace{-6mm}
\caption{Pre-training bias results in Data Wrangler.\label{fig:wrangler}}
\end{center}
\end{figure}

For illustration we present a simple case study with the
 German Credit Data dataset from the UCI repository~\cite{Dua:2019German} that contains \num{1000} labeled credit applications. Features include credit purpose, credit amount, housing status, and employment history. Categorical attributes are converted to indicator variables, e.g., A30 means ``no credits taken.'' We mapped the labels (Class1Good2Bad) to 0 for poor credit and 1 for good.

{\noindent \bf Bias analysis.}
Before we train the model we seek to explore biases in the dataset. Figure~\ref{fig:wrangler} shows \clarify integrated into Data Wrangler. Convenient drop-downs are available to select the label column and the group. Also, by default a small set of metrics are pre-selected in order to not overwhelm customers with too many metrics. No code needs to be written to conduct this analysis.
From the class imbalance value of -0.94, we infer that the dataset is very imbalanced, with most examples representing foreign workers.
However, as $DPL$ is close to $0$, there is not much difference in the label distributions across the two groups. 
\begin{figure}
\vspace{-4mm}
\begin{center}
  \includegraphics[width=1\linewidth]{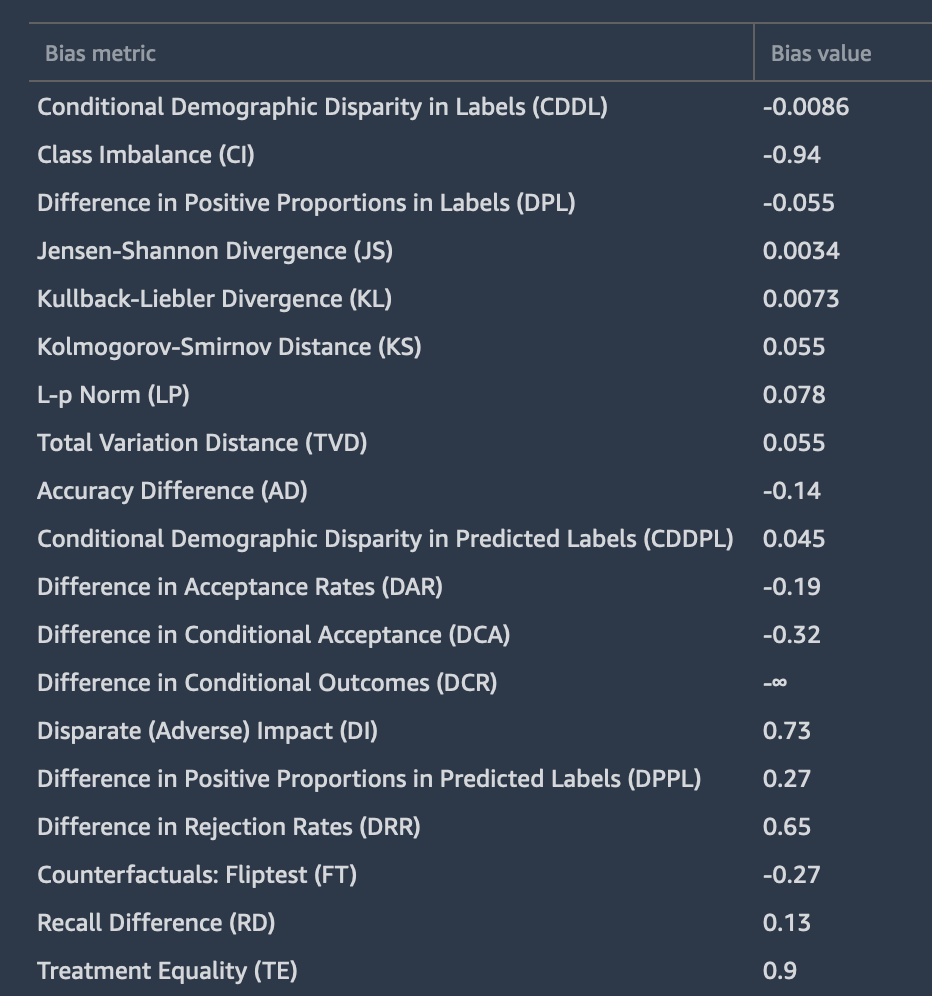}
\caption{Post-training bias results in Studio Trials.\label{fig:postbias}}
\end{center}
\end{figure}
\begin{figure}
\begin{center}
  \includegraphics[width=1\linewidth]{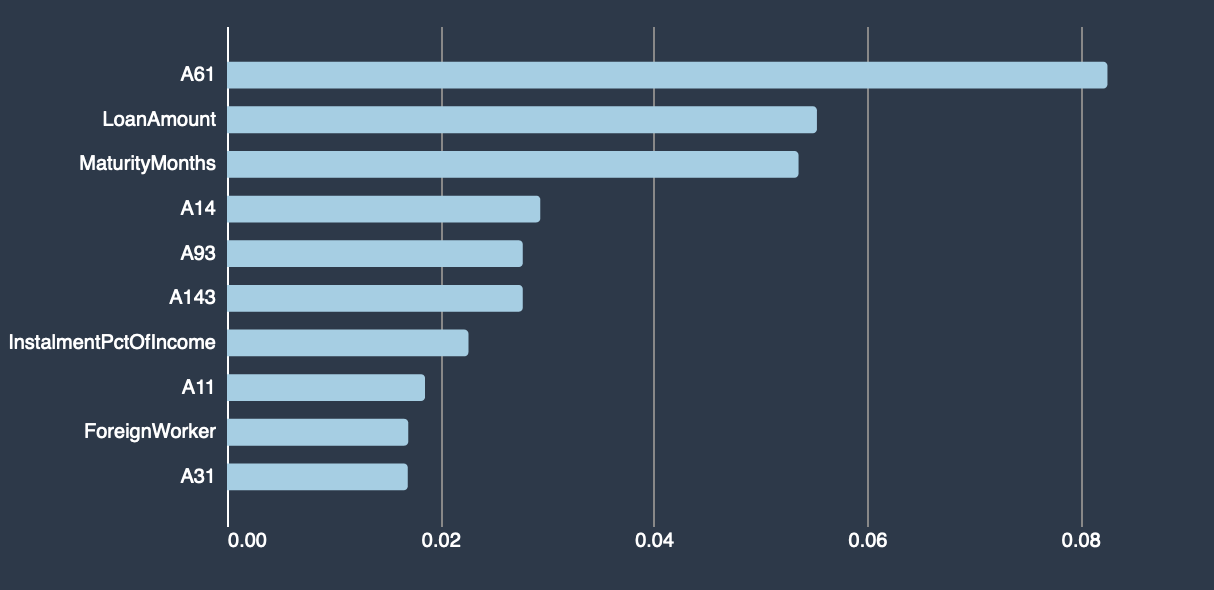}
\vspace{-3mm}
\caption{SHAP feature importance.\label{fig:shap}}
\end{center}
\end{figure}
With these encouraging results we proceed with the training of an XGBoost model.
We submit the configuration for the post-training bias analysis (see Appendix~\ref{sec:appendix_config}) as input along with the validation data to the \clarify  job. 
The results shown in Figure~\ref{fig:postbias} reveal a mixed picture with positive and negative metrics. We start by inspecting the disparate impact $DI$  with a value of 0.73 much below 1. Since the label rates are almost the same across the two groups ($DPL$), $DI$ is driven by differences in precision and recall. We see that foreign workers have a lower recall ($RD$ of 0.13), but a higher precision ($DAR$ of -0.19). Corrective steps can mitigate this bias prior to deployment.

{\noindent \bf Feature importance.}
To understand which features are affecting the model's predictions the most we look at the global SHAP values that are averaged across all examples after taking the absolute value.
The three most important features are A61 (having very little savings), the loan amount, and credit duration, see Figure~\ref{fig:shap}.

\section{Deployment Results}\label{sec:results}
\begin{table}
    \centering
    \begin{tabular}{r | c | c || c }
    \toprule
     & evaluation & monitoring & overall \\
    \midrule
    bias & 0.15 & 0.21 & 0.36\\
    explainability & 0.43 & 0.21 & 0.64 \\
    \midrule
    overall  & 0.58 & 0.42 & 1 
    \end{tabular}
    \caption{Break-down of \clarify jobs across method (bias / explainability) and phase (evaluation / monitoring).}
    \label{tab:breakdown}
    \vspace{-4mm}
\end{table}
In the first few months after launch 36\% of all \clarify jobs computed bias metrics while 64\% computed feature importance, see Table~\ref{tab:breakdown}. Monitoring jobs made up 42\% of jobs, illustrating that customers reached a mature stage in the ML lifecycle in which they incorporated bias and feature importance monitoring. This is expected in a healthy development cycle; after a few iterations on model training and evaluation, a model passes the launch criteria and is continuously monitored after. 

Next, we evaluate our design goals and highlight two use cases.

\subsection{Wide Applicability}
Customers across several industry verticals
use \clarify, including digital publishing, education, electronics, finance, government, hardware, healthcare, IT, manufacturing, news, sports, staffing, and telecommunications.
Customers use \clarify for binary predictions, multi-class predictions, and ranking tasks. They are located in USA, Japan, UK, and Norway among other countries. We present two specific customer use cases in \S\ref{sec:customers}.

\subsection{Scalability}\label{sec:scalability}

\begin{figure*}
\vspace{-6mm}
\begin{subfigure}{.5\textwidth}
  \centering
  \includegraphics[width=.9\linewidth]{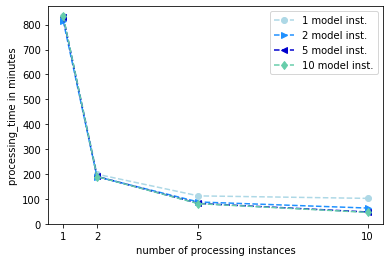}
\end{subfigure}%
\begin{subfigure}{.5\textwidth}
  \centering
   \includegraphics[width=.9\linewidth]{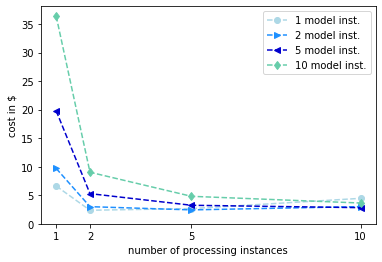}
\end{subfigure}
\vspace{-4mm}
\caption{Scaling of SHAP on \num{100000} examples and measuring time (left) and cost (right) for a varying number of instances. \label{fig:shap_case_of_study}}
\label{fig:fig}
\vspace{-4mm}
\end{figure*}

We conducted experiments measuring the time taken to compute bias metrics and feature importance scores. We oversampled the German credit data set described in \S\ref{sec:case_study} %
creating up 
to 1 million examples to study scalability. We ran experiments on {\tt ml.c5.xlarge} machines which are suitable for this  workload and provide cost estimates based on AWS pricing in the {\tt us-west-2} region. 

{\noindent \bf Bias metrics.}
\begin{table}[]
    \centering
    \begin{tabular}{l| c | c}
    \toprule
    method & processing time in minutes & cost in dollars \\
    \midrule
      pre-training bias   & 1  & \$0.03\\
      post-training bias   & 14 & \$0.13
    \end{tabular}
    \caption{Processing time and cost of bias metrics computation for 1 million examples.}
    \label{tab:bias_results}
\vspace{-4mm}
\end{table}
Table~\ref{tab:bias_results} shows that the bias metrics computation is fast even for 1M examples. Post-training bias metrics are considerably slower than pre-training metrics since they require model inference. %
The reported time includes 3-5 minutes needed to spin up the model endpoint. %
Scalability issues may arise when the dataset does not fit in memory %
(in which case a larger instance type can be chosen) or when model inference is much slower (in which case  the instance count can be increased).

{\noindent \bf Explainability.}
\begin{table}[]
    \centering
    \begin{tabular}{l| c | c}
    \toprule
    number of examples & processing time in minutes & cost in dollars \\
    \midrule
      1000  & 16  & \$0.13\\
      \num{10000}   & 90 & \$0.71 \\
       \num{100000}   & 811 & \$6.43
    \end{tabular}
    \caption{Processing time and cost of  SHAP with varying  dataset sizes using one processing / model instance.}
    \label{tab:shap}
    \vspace{-6mm}
\end{table}
Table~\ref{tab:shap} shows that time and cost scale roughly linearly with the dataset size. The implementation of SHAP iterates over all examples in a for-loop and the execution at each step  dominates the time and cost.
For \num{100000} examples the running time of more than 13h is so long that it can be a point of frustration for customers. We parallelized SHAP to reduce the running time.%

Figure~\ref{fig:shap_case_of_study} shows the scaling behavior of SHAP on \num{100000} examples. We see that increasing the number of instances (both for the processing job and the model endpoint) to two %
reduces 
the 
time down to 3h and 17 minutes. This super-linear speedup is surprising. It is due to the fact that our single instance implementation does not fully utilize all cores. Our Spark implementation not only splits the workload across machines but also across cores within those machines thereby reducing not just time but even cost.
We observe that 1 model endpoint instance for every 1 to 2 processing instances provides a good ratio: 
A higher ratio underutilizes model endpoint instances, increasing cost without a speed up. With a lower ratio (e.g., 1 model endpoint instance for 10 processing instances), the model becomes the bottleneck, and cost as well as running time can be reduced by increasing the number of model endpoint instances.
\begin{table}[]
    \centering
    \begin{tabular}{l| c | c | c }
    \toprule
    method & \multicolumn{3}{c}{time in minutes}  \\
    & P50 & P90 & P99 \\
    \midrule
      bias  &  1 &	8 &	11 \\
      explainability &	5 &	12	& 103 \\
      \midrule
      overall  & 4 & 12 & 101 \\
    \end{tabular}
    \caption{Percentiles of processing times in minutes.}
    \label{tab:running_times}
    \vspace{-8mm}
\end{table}

{\noindent \bf Customer results.}
Half of the \clarify jobs of  customers take at most 4 minutes. Table~\ref{tab:running_times} shows percentiles of processing times for bias as well as explainability jobs. We see that the running time is less than two hours for 99\% of the jobs.
Across all jobs 35\% use multiple processing instances thereby reducing running time. At most 10 instances were used by customers so far.

\subsection{Ease of use}\label{sec:usability}
We obtained qualitative customer feedback on usability. 
For customers it was not always easy to get the configuration right despite documentation. Difficulties included setting the right fields with the right values, and figuring out fields that are optional, or need to go together for a specific use case.
These difficulties were reduced through tooling, documentation, and examples. 
In terms of tooling, customers can use  the explanatory UI or programmatically create the configuration. 
Regarding documentation, customers requested additional documentation covering specifics of multi-class prediction.
Regarding examples, multiple customers with strict security requirements who used advanced configuration options to enable encryption of data at rest and in transit with fine-grained access control benefited from \clarify examples for such settings.
Other customers noted that our notebooks focus on the CSV format and requested examples with the JSONLines data format. 
These examples illustrate tension between the goals of wide applicability and usability: Ease of use for one setting (e.g., CSV data, binary predictions) does not extend to all settings (e.g., JSONLines data, multi-class prediction).

We received several inquiries about formulating the development, business, and regulatory needs into the framework that \clarify provides. 
While we were able to find a fit for many different use cases, these questions are hard and answers differ case-by-case. 
For example, the question of which bias metric is important, depends on the application. There is no way to get around this question, as it is impossible to remove all biases~\cite{Chouldechova2017FairPW, kleinberg2016inherent}. Here again, documentation helps customers to address this hard question. %

{\noindent \bf Quantitative customer results.} 
SageMaker AutoPilot, that automatically builds, trains, and tunes models,  integrated \clarify in March 2021. AutoPilot contributes 54\% of the explainability jobs. 
The automatic configuration functionality offered by AutoPilot, along with the other improvements listed above, helped reduce failure rates by 75\%.

\subsection{Two Customer Use Case Examples}\label{sec:customers}
Next, we illustrate the use of \clarify in production by providing two real customer use cases.

{\noindent \bf Fraud detection.}
Zopa's data scientists use several dozens of input features such as application details, device information, interaction behaviour, and demographics. For model building, they extract the training dataset from their Amazon Redshift data warehouse, perform data cleaning, and store into Amazon S3. As Zopa has their own in-house ML library for both feature engineering and ML framework support, they use the ``Bring Your Own Container'' (BYOC) approach to leverage SageMaker’s managed services and advanced functionalities such as hyperparameter optimization. The optimized models are then deployed through a Jenkins CI/CD pipeline to the production system and serve as a microservice for real-time fraud detection as part of their customer facing platform.

Explanations are obtained both during model training for validation, and after deployment for model monitoring and generating insights for underwriters. These are done in a non-customer facing analytical environment due to a heavy computational requirement and a high tolerance of latency. SageMaker Multi Model Server stack is used in a similar BYOC fashion, to register the models for the \clarify processing job. \clarify spins up an ephemeral model endpoint and invokes it for millions of predictions on synthetic contrastive data. These predictions are used to compute SHAP values for each individual example, which are stored in S3. 

For model validation and monitoring,  \clarify aggregates these local feature importance values across examples in the training/monitoring data, to obtain global feature importance, and visualizes them in SageMaker Studio. 
To give insights to operation, data scientists select  features with the highest SHAP values that contributed most to a positive fraud score (i.e., likely fraudster) for each individual example, and report them to the underwriter. %

{\noindent \bf Explanation of a soccer scoring probability model.}
In the context of soccer matches the DFL (Deutsche Fußball Liga) applied \clarify to their xGoal model. The model predicts the likelihood that an attempted shot will result in a goal. 
They found that overall the features {\it AngleToGoal} and {\it DistanceToGoal} have the highest importance across all examples. 
Let us consider one of the most interesting games of the 2019/2020 Bundesliga season, where Borussia Dortmund beat Augsburg in a 5-3 thriller on January 18, 2020.
 Taking a look at the sixth goal of the game, scored by Jadon Sancho, the model predicted the goal with a high xGoal prediction of 0.18 compared to the average of prediction  of 0.0871 across the past three seasons. What explains this high prediction? The SHAP analysis reveals that {\it PressureSum}, {\it DistanceToGoalClosest}, and {\it AmountofPlayers} contribute the most to this prediction. This is a noticeable deviation from the features that are globally the most important.
 What makes this explanation convincing is the unusual nature of this goal: Sancho received a highly offensive through ball well into Augsburg’s half (past all defenders – hence low {\it PressureSum}), only to dart round the keeper and tap it into the goal near the posts (low {\it AngleToGoal}).

\section{Lessons Learned In Practice}~\label{sec:lessons}

{\noindent \bf Wide Applicability}:
We learned that wide applicability across datasets and model endpoints not only requires flexible parsing logic (see \S\ref{sec:development}), but also creates a tension with ease of use:
Our code tries to infer whether a file contains a header, or a feature is categorical or numerical. While it is convenient for most not having to specify all these details, for some customers our inference does not work well. %
This tension is further explored in \S\ref{sec:usability}.

{\noindent \bf Scalability}: When working towards our goal of creating a scalable application, we learned that bias and explainability analyses have different resource demands (SHAP can be compute intensive, while bias metrics can be memory intensive), and hence require separate tuning of the parallelism.
For explainability we had to work on small chunks, while for bias reports, we had to work on large chunks of the dataset. For details and other technical lessons, see \S\ref{sec:dev_challenges}.

{\noindent \bf Usability}: In retrospect, our biggest lesson concerning usability is one we should have recognized prior to the usability studies -- we cannot rely heavily on documentation. Instead we need to enable users to try out the feature without reading documentation.
Early implementations of \clarify required the user to create their JSON configuration file manually. This was prone to both schema errors caused by users creating malformed JSON, and soft errors caused by users defining metrics using incorrect fields or labels. To avoid such errors we developed an explanatory UI.
Now, we ask the customer to ``Select the column to analyze for bias'' and provide a drop down menu or programmatically construct the configuration for them. We are further revising the UI, see \S\ref{sec:usability}.%

{\noindent \bf Best Practices}: We recognize that the notions of bias and fairness are highly dependent on the application and that the choice of the attribute(s) for which to measure bias, as well as the choice of the bias metrics, can be guided by social, legal, and other non-technical considerations. Our experience suggests that the successful adoption of fairness-aware ML approaches in practice requires building consensus and achieving collaboration across key stakeholders (such as product, policy, legal, engineering, and AI/ML teams, as well as end users and communities). Further, fairness and explainability related ethical considerations need to be taken into account during each stage of the ML lifecycle, for example, by asking questions stated in Figure~\ref{fig:cycle}.

\section{Conclusion}\label{sec:conclusions}
Motivated by the need for bias detection and explainability from regulatory, business, and data science perspectives, we presented Amazon SageMaker Clarify, a new functionality 
that helps detect bias in data and models and helps explain predictions made by models. We described the design considerations, methodology, and technical architecture of \clarify, as well as how it is integrated across the ML lifecycle (data preparation, model evaluation, and monitoring post deployment). We discussed key technical challenges and resolutions, a case study, customer use cases, deployment results, and lessons learned in practice. Considering that \clarify is a scalable, cloud-based bias and explainability service designed to address the needs of customers from multiple industries, the insights and experience from our work are likely to be of interest for researchers and practitioners working on responsible AI systems.

While we have focused on providing tools for bias and explainability in the context of an ML lifecycle, it is important to realize that this is an ongoing effort that is part of a larger process.
Another key challenge is to navigate the trade-offs in the context of an application between model accuracy, interpretability, various bias metrics, and related dimensions such as privacy and robustness (see, for example, ~\cite{Chouldechova2017FairPW, kleinberg2016inherent,kleinberg-interpretability-fairness,reconstruction:fatMilliSDH19}).
It cannot be overemphasized that developing AI solutions needs to be thought of more broadly as a process involving iterated inputs from and discussions with key stakeholders such as product, policy, legal, engineering, and AI/ML teams as well as end users and communities, and asking relevant questions during all stages of the ML lifecycle.

\bibliographystyle{abbrv}

% \bibliographystyle{abbrv}
% %\bibliography{references}

\begin{thebibliography}{10}

\bibitem{wh2014bigdata}
Big data: Seizing opportunities, preserving values.
\newblock
  \url{https://obamawhitehouse.archives.gov/sites/default/files/docs/big_data_privacy_report_may_1_2014.pdf},
  2014.

\bibitem{wh2016bigdata}
Big data: A report on algorithmic systems, opportunity, and civil rights.
\newblock
  \url{https://obamawhitehouse.archives.gov/sites/default/files/microsites/ostp/2016_0504_data_discrimination.pdf},
  2016.

\bibitem{ftc2016bigdata}
Big data: A tool for inclusion or exclusion? {Understanding} the issues ({FTC}
  report).
\newblock
  \url{https://www.ftc.gov/reports/big-data-tool-inclusion-or-exclusion-understanding-issues-ftc-report},
  2016.

\bibitem{agarwal_reductions_2018}
A.~Agarwal, A.~Beygelzimer, M.~Dudik, J.~Langford, and H.~Wallach.
\newblock A reductions approach to fair classification.
\newblock In {\em ICML}, 2018.

\bibitem{barocas_fairness_2019}
S.~Barocas, M.~Hardt, and A.~Narayanan.
\newblock {\em Fairness and {Machine} {Learning}}.
\newblock fairmlbook.org, 2019.

\bibitem{barocas_hidden_2020}
S.~Barocas, A.~D. Selbst, and M.~Raghavan.
\newblock The hidden assumptions behind counterfactual explanations and
  principal reasons.
\newblock In {\em {FAccT}}, 2020.

\bibitem{berk_fairness_2017}
R.~Berk, H.~Heidari, S.~Jabbari, M.~Kearns, and A.~Roth.
\newblock Fairness in criminal justice risk assessments: The state of the art.
\newblock {\em Sociological Methods \& Research}, 2018.

\bibitem{bhatt2020explainable}
U.~Bhatt, A.~Xiang, S.~Sharma, A.~Weller, A.~Taly, Y.~Jia, J.~Ghosh, R.~Puri,
  J.~M. Moura, and P.~Eckersley.
\newblock Explainable machine learning in deployment.
\newblock In {\em FAccT}, 2020.

\bibitem{black_fliptest_2020}
E.~Black, S.~Yeom, and M.~Fredrikson.
\newblock {FlipTest}: Fairness testing via optimal transport.
\newblock In {\em FAccT}, 2020.

\bibitem{calders_three_2010}
T.~Calders and S.~Verwer.
\newblock Three naive {Bayes} approaches for discrimination-free
  classification.
\newblock {\em Data Mining and Knowledge Discovery}, 21(2), Sept. 2010.

\bibitem{Chouldechova2017FairPW}
A.~Chouldechova.
\newblock Fair prediction with disparate impact: A study of bias in recidivism
  prediction instruments.
\newblock {\em Big data}, 5 2, 2017.

\bibitem{donini_empirical_2018}
M.~Donini, L.~Oneto, S.~Ben-David, J.~Shawe-Taylor, and M.~Pontil.
\newblock Empirical risk minimization under fairness constraints.
\newblock In {\em NeurIPS}, 2018.

\bibitem{Dua:2019German}
D.~Dua and C.~Graff.
\newblock {UCI} machine learning repository, 2017.

\bibitem{dwork-fairness}
C.~Dwork, M.~Hardt, T.~Pitassi, O.~Reingold, and R.~Zemel.
\newblock Fairness through awareness.
\newblock In {\em ITCS}, 2012.

\bibitem{efron1994introduction}
B.~Efron and R.~J. Tibshirani.
\newblock {\em An introduction to the bootstrap}.
\newblock CRC press, 1994.

\bibitem{geyik2019fairness}
S.~C. Geyik, S.~Ambler, and K.~Kenthapadi.
\newblock Fairness-aware ranking in search \& recommendation systems with
  application to {LinkedIn} talent search.
\newblock In {\em KDD}, 2019.

\bibitem{goodman2017european}
B.~Goodman and S.~Flaxman.
\newblock European union regulations on algorithmic decision-making and a
  “right to explanation”.
\newblock {\em AI magazine}, 2017.

\bibitem{guidotti2018survey}
R.~Guidotti, A.~Monreale, S.~Ruggieri, F.~Turini, F.~Giannotti, and
  D.~Pedreschi.
\newblock A survey of methods for explaining black box models.
\newblock {\em ACM Computing Surveys}, 51(5), 2018.

\bibitem{hardt_equality_2016}
M.~Hardt, E.~Price, E.~Price, and N.~Srebro.
\newblock Equality of opportunity in supervised learning.
\newblock In {\em NeurIPS}, 2016.

\bibitem{holstein2019improving}
K.~Holstein, J.~Wortman~Vaughan, H.~Daum{\'e}~III, M.~Dudik, and H.~Wallach.
\newblock Improving fairness in machine learning systems: What do industry
  practitioners need?
\newblock In {\em CHI}, 2019.

\bibitem{janzing_feature_2019}
D.~Janzing, L.~Minorics, and P.~Bloebaum.
\newblock Feature relevance quantification in explainable {AI}: A causal
  problem.
\newblock In {\em AISTATS}, 2020.

\bibitem{causal_fair}
N.~Kilbertus, M.~Rojas~Carulla, G.~Parascandolo, M.~Hardt, D.~Janzing, and
  B.~Sch\"{o}lkopf.
\newblock Avoiding discrimination through causal reasoning.
\newblock In {\em NeurIPS}, 2017.

\bibitem{kleinberg-interpretability-fairness}
J.~Kleinberg and S.~Mullainathan.
\newblock Simplicity creates inequity: Implications for fairness, stereotypes,
  and interpretability.
\newblock In {\em EC}, 2019.

\bibitem{kleinberg2016inherent}
J.~Kleinberg, S.~Mullainathan, and M.~Raghavan.
\newblock Inherent trade-offs in the fair determination of risk scores.
\newblock In {\em ITCS}, 2017.

\bibitem{counterfactual_failr}
M.~J. Kusner, J.~Loftus, C.~Russell, and R.~Silva.
\newblock Counterfactual fairness.
\newblock In {\em NeurIPS}, 2017.

\bibitem{lakkaraju_faithful_2019}
H.~Lakkaraju, E.~Kamar, R.~Caruana, and J.~Leskovec.
\newblock Faithful and customizable explanations of black box models.
\newblock In {\em AIES}, 2019.

\bibitem{sagemaker}
E.~Liberty, Z.~Karnin, B.~Xiang, L.~Rouesnel, B.~Coskun, R.~Nallapati,
  J.~Delgado, A.~Sadoughi, Y.~Astashonok, P.~Das, C.~Balioglu, S.~Chakravarty,
  M.~Jha, P.~Gautier, D.~Arpin, T.~Januschowski, V.~Flunkert, Y.~Wang,
  J.~Gasthaus, L.~Stella, S.~Rangapuram, D.~Salinas, S.~Schelter, and A.~Smola.
\newblock Elastic machine learning algorithms in {Amazon SageMaker}.
\newblock In {\em SIGMOD}, 2020.

\bibitem{mythos}
Z.~C. Lipton.
\newblock The mythos of model interpretability.
\newblock {\em ACM Queue}, 16(3), 2018.

\bibitem{lundberg_unified_2017}
S.~M. Lundberg and S.-I. Lee.
\newblock A unified approach to interpreting model predictions.
\newblock In {\em NeurIPS}, 2017.

\bibitem{lundberg_explainable_2018}
S.~M. Lundberg, B.~Nair, M.~S. Vavilala, M.~Horibe, M.~J. Eisses, T.~Adams,
  D.~E. Liston, D.~K.-W. Low, S.-F. Newman, J.~Kim, and S.-I. Lee.
\newblock Explainable machine-learning predictions for the prevention of
  hypoxaemia during surgery.
\newblock {\em Nature Biomedical Engineering}, 2(10), Oct. 2018.

\bibitem{mehrabi_survey_2019}
N.~Mehrabi, F.~Morstatter, N.~Saxena, K.~Lerman, and A.~Galstyan.
\newblock A survey on bias and fairness in machine learning.
\newblock {\em arXiv preprint arXiv:1908.09635}, 2019.

\bibitem{reconstruction:fatMilliSDH19}
S.~Milli, L.~Schmidt, A.~D. Dragan, and M.~Hardt.
\newblock Model reconstruction from model explanations.
\newblock In {\em FAccT}, 2019.

\bibitem{mitchell2019model}
M.~Mitchell, S.~Wu, A.~Zaldivar, P.~Barnes, L.~Vasserman, B.~Hutchinson,
  E.~Spitzer, I.~D. Raji, and T.~Gebru.
\newblock Model cards for model reporting.
\newblock In {\em FAccT}, 2019.

\bibitem{pearl_causal_2009}
J.~Pearl.
\newblock Causal inference in statistics: {An} overview.
\newblock {\em Statistics Surveys}, 3, 2009.

\bibitem{perrone2020fair}
V.~Perrone, M.~Donini, M.~B. Zafar, R.~Schmucker, K.~Kenthapadi, and
  C.~Archambeau.
\newblock Fair {Bayesian} optimization.
\newblock In {\em AIES}, 2021.

\bibitem{ribeiro2016should}
M.~T. Ribeiro, S.~Singh, and C.~Guestrin.
\newblock "{Why} should {I} trust you?" {Explaining} the predictions of any
  classifier.
\newblock In {\em KDD}, 2016.

\bibitem{rudin_stop_2019}
C.~Rudin.
\newblock Stop explaining black box machine learning models for high stakes
  decisions and use interpretable models instead.
\newblock {\em Nature Machine Intelligence}, 1(5), May 2019.

\bibitem{shapley_value_1952}
L.~S. Shapley.
\newblock A value for n-person games.
\newblock {\em Contributions to the Theory of Games}, 2(28), 1953.

\bibitem{NEURIPS2019_0e1feae5}
S.~Sharifi-Malvajerdi, M.~Kearns, and A.~Roth.
\newblock Average individual fairness: Algorithms, generalization and
  experiments.
\newblock In {\em NeurIPS}, 2019.

\bibitem{simonyan_deep_2014}
K.~Simonyan, A.~Vedaldi, and A.~Zisserman.
\newblock Deep inside convolutional networks: Visualising image classification
  models and saliency maps.
\newblock In {\em ICLR Workshop}, 2014.

\bibitem{rankingfair}
A.~Singh and T.~Joachims.
\newblock Fairness of exposure in rankings.
\newblock In {\em KDD}, 2018.

\bibitem{slack2021defuse}
D.~Slack, N.~Rauschmayr, and K.~Kenthapadi.
\newblock Defuse: Harnessing unrestricted adversarial examples for debugging
  models beyond test accuracy.
\newblock {\em arXiv preprint arXiv:2102.06162}, 2021.

\bibitem{sundararajan_axiomatic_2017}
M.~Sundararajan, A.~Taly, and Q.~Yan.
\newblock Axiomatic attribution for deep networks.
\newblock In {\em ICML}, 2017.

\bibitem{vasudevan20lift}
S.~Vasudevan and K.~Kenthapadi.
\newblock {LiFT}: A scalable framework for measuring fairness in {ML}
  applications.
\newblock In {\em CIKM}, 2020.

\bibitem{wachter_why_2020}
S.~Wachter, B.~Mittelstadt, and C.~Russell.
\newblock Why fairness cannot be automated: Bridging the gap between {EU}
  non-discrimination law and {AI}.
\newblock {SSRN} {Scholarly} {Paper} ID 3547922, Social Science Research
  Network, Mar. 2020.

\bibitem{what-if}
J.~{Wexler}, M.~{Pushkarna}, T.~{Bolukbasi}, M.~{Wattenberg}, F.~{Viégas}, and
  J.~{Wilson}.
\newblock The {What-If Tool}: Interactive probing of machine learning models.
\newblock {\em IEEE Transactions on Visualization and Computer Graphics},
  26(1), 2020.

\bibitem{fair-classification}
M.~B. Zafar, I.~Valera, M.~Gomez~Rodriguez, and K.~P. Gummadi.
\newblock Fairness beyond disparate treatment \& disparate impact: Learning
  classification without disparate mistreatment.
\newblock In {\em WWW}, 2017.

\bibitem{adv_fairness}
B.~H. Zhang, B.~Lemoine, and M.~Mitchell.
\newblock Mitigating unwanted biases with adversarial learning.
\newblock In {\em AIES}, 2018.

\end{thebibliography}

\newpage
\begin{acks}
We would like to thank our colleagues at AWS for their contributions to \clarify. 
We thank Edi Razum, Luuk Figdor, Hasan Poonawala, the DFL and Zopa for contributions to the customer use case, and C{\'e}dric Archambeau and the anonymous reviewers for their valuable feedback.

\end{acks}

\appendix

\section{Detailed Configuration}\label{sec:appendix_config}

\subsection{Additional Bias Metrics}\label{sec:appendix_bias_methods}
Recall that we defined the fraction of positively labeled examples in group $a$ as $q_a=n_a^{(1)}/n_a$ ($q_d=n_d^{(1)}/n_d$ for group $d$, respectively).
They define a probability distribution $P_a$ for group $a$ over labels as $q_a$ for positive labels and $1-q_a$ for negative labels and $P_d$ as $q_d$ and $1-q_d$. 
Differences in these probability distributions yield additional pre-training bias metrics:

\begin{enumerate}[leftmargin=*]
\item[$KL$] %
KL divergence measures how much information is needed to move from one probability distribution  $P_a$ to another  $P_d$: $KL = q_a \log \frac{q_a}{q_d} + (1-q_a) \log \frac{1-q_a}{1-q_d}$. 

\item[$JS$] Jenson-Shannon divergence provides a symmetric form if KL divergence. Denoting by $P$ the average of the label distributions of the two groups, we define
$$
\textstyle
JS(P_a,P_d) = \frac{1}{2}\left[KL(P_a,P) + KL(P_d,P)\right].
$$

\item[$LP$] We can also consider the norm between the probability distributions of the labels for the two groups. For $p\geq1$, we have
$$
\textstyle
L_p(P_a,P_d) =  \left[\sum_y |P_a(y)-P_d(y)|^p \right]^{1/p}.
$$

\item[$TVD$] The total variation distance is half the $L_1$ distance: $TVD = \frac{1}{2}L_1(P_a, P_d)$.

\item[$KS$] Kolmogorov-Smirnov considers the maximum difference across label values: $KS = \max(|P_a - P_d|)$.
\end{enumerate}

An additional post-training bias metric is based on a flip test.
\begin{enumerate}[leftmargin=*]
\item[$FT$] Counterfactual Fliptest: $FT$ assesses whether similar examples across the groups receive similar predictions. For each example in $d$, we count how often its k-nearest neighbors in $a$ receive a different prediction. 
Simplifying \cite{black_fliptest_2020}, we define
$FT = (F^{+}-F^{-}) / n_d$
where $F^+$ ($F^-$ resp.) is the number of examples in  $d$ with a negative (positive, resp.) prediction whose nearest neighbors in   $a$ have a positive (negative, resp.) prediction.
\end{enumerate}

\subsection{Configuration Options}\label{sec:appendix_predictor_config}
Table~\ref{tab:clarify_config} illustrates the flexible support for model outputs.
\begin{table}[h!]
    \centering
    \begin{tabular}{c | l }
    \toprule
    Example output & Configuration of predictor \\
    \midrule
      {\tt C3},"[0.1,0.3,0.6]"  & {\tt "label": 0,} \\
      & {\tt "probability": 1} \\
      \midrule
      $"[0.1,0.3,0.6]"$  & {\tt "probability": 0,} \\
      & ${\tt "label\_headers"}: [{\tt "C1"}, {\tt"C2"}, {\tt"C3"}]$ \\
      \midrule
      $\{"{\tt pred}": {\tt C3},$ & {\tt "label": "pred",} \\
  $"{\tt sc}":[0.1,0.3,0.6]\}$ & {\tt "probability": "sc",} \\
           & {\tt "content\_type"}: {\tt \small "application/jsonlines"}
    \end{tabular}
    \caption{Illustration of multi-class model outputs supported by \clarify. The model predicts scores for classes {\tt C1, C2, C3}.}
    \label{tab:clarify_config}
    \vspace{-4mm}
\end{table}

\subsection{Detailed Example}\label{sec:appendix_ex_config}
Below we show an example analysis config. It is minimal and additional configuration options (e.g. to support datasets without headers or model endpoints using JSON format) exist. 

\begin{minted}[
    gobble=4,
    frame=single,
    linenos
  ]{yaml}
{
    "dataset_type": "text/csv",
    "label": "Class1Good2Bad",
    "label_values_or_threshold": [1],
    "facet": [
        {
         "name_or_index" : "ForeignWorker",
         "value_or_threshold": [1]
        }
    ],
    "group_variable": "A151",
    "methods": {
        "shap": {
            "baseline": "s3://<bucket>/<prefix>/<file>",
            "num_samples": 3000,
            "agg_method": "mean_abs"
        },
        "pre_training_bias": {
            "methods": "all"
        },
        "post_training_bias": {
            "methods": "all"
        },
        "report": {
            "name": "report",
            "title": "Clarify Analysis Report"
        }
    },
    "predictor": {
        "model_name": "german-xgb",
        "instance_type": "ml.c5.xlarge",
        "initial_instance_count": 1
    }
}
\end{minted}

\end{document}